\documentclass{article} 
\usepackage{iclr2025_conference,times}

\usepackage{amsmath,amsfonts,bm}

\def\eqref#1{equation~\ref{#1}}

\def\1{\bm{1}}

\DeclareMathAlphabet{\mathsfit}{\encodingdefault}{\sfdefault}{m}{sl}
\SetMathAlphabet{\mathsfit}{bold}{\encodingdefault}{\sfdefault}{bx}{n}

\usepackage{hyperref}
\usepackage{url}
\usepackage{booktabs}
\usepackage{multirow} 
\usepackage{subcaption}
\usepackage{arydshln}
\usepackage{todonotes}
\usepackage{utfsym}
\usepackage{algorithm}
\usepackage{algpseudocode}
\usepackage{amsmath}
\usepackage{setspace}
\usepackage{wrapfig} 
\usepackage{colortbl}
\usepackage{makecell}
\usepackage{tcolorbox}
\usepackage{diagbox}
\usepackage{arydshln}

\tcbuselibrary{most}

\definecolor{lightgreen}{RGB}{200,255,200}
\definecolor{lightpink}{rgb}{1.0, 0.85, 0.9} 
\definecolor{lightblue}{rgb}{0.529, 0.808, 0.922} 
\definecolor{lightgray}{gray}{0.85}

\newtcolorbox{mybox}[2][]
  {colback = black!5!white, colframe = black!75!black, fonttitle = \bfseries,
    colbacktitle = black!100!black, enhanced,
    attach boxed title to top left={yshift=-2.2mm,xshift=4mm},
    title=#2,#1}

\title{Unlocking Efficient Long-to-Short LLM Reasoning with Model Merging}

\author{Han Wu$^*$, Yuxuan Yao$^*$, Shuqi Liu$^*$, Zehua Liu$^*$, Xiaojin Fu, Xiongwei Han, Xing Li\\\textbf{Hui-Ling Zhen, Tao Zhong, Mingxuan Yuan} \\ \\
\quad\quad\quad\quad\quad\quad\quad\quad\quad\quad\quad\quad\quad~~~~ \textbf{Huawei Noah's Ark Lab} \\ \\
\texttt{\{wu.han1,hanxiongwei,zhongtao5,Yuan.Mingxuan\}@huawei.com}
}

\iclrfinalcopy 
\begin{document}

\maketitle

\begin{figure}[h]
    \centering
    \begin{subfigure}[b]{0.48\textwidth}
        \centering
        \includegraphics[width=\textwidth]{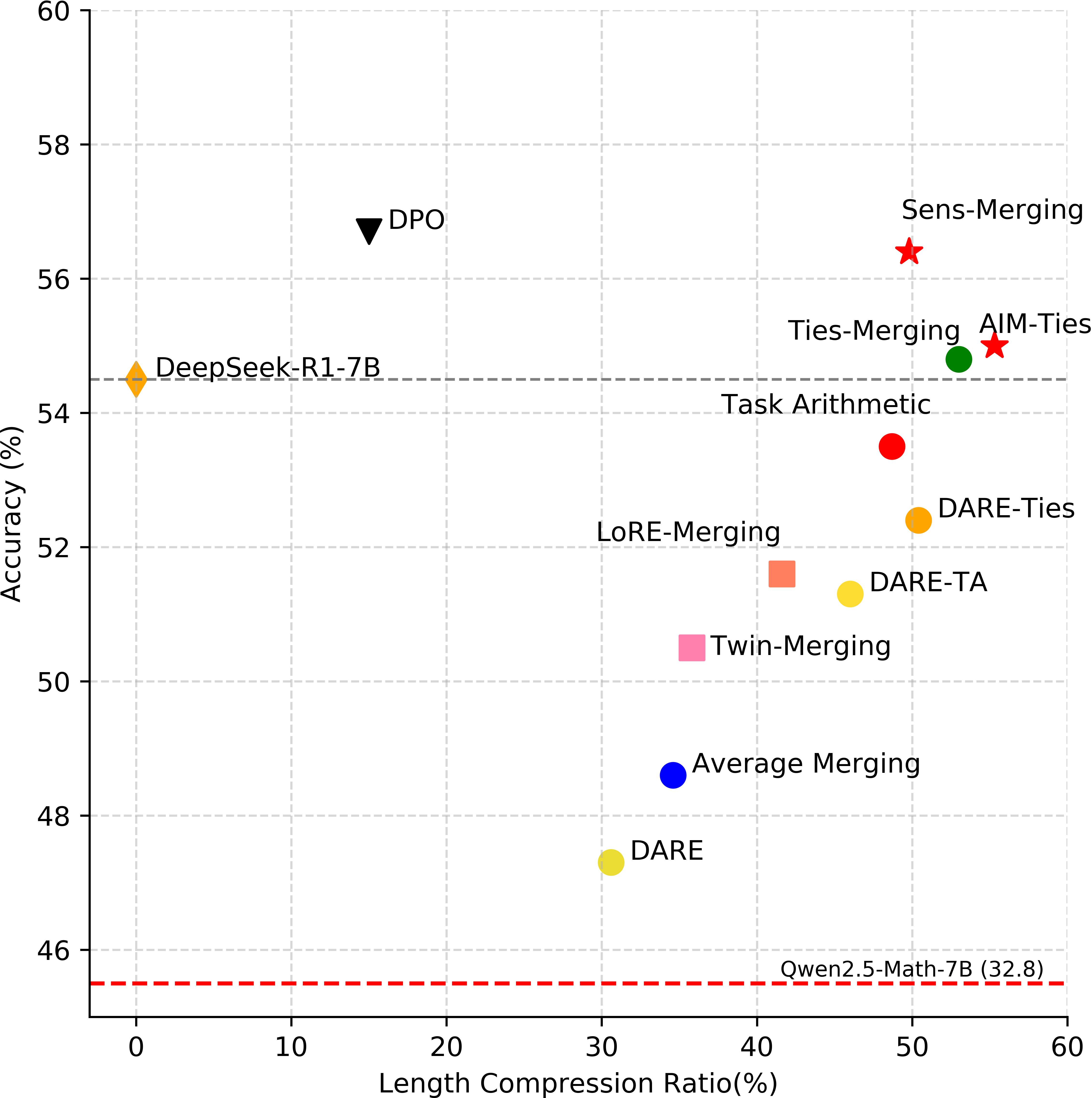}
    \end{subfigure}
    \hfill
    \begin{subfigure}[b]{0.46\textwidth}
        \centering
        \includegraphics[width=\textwidth]{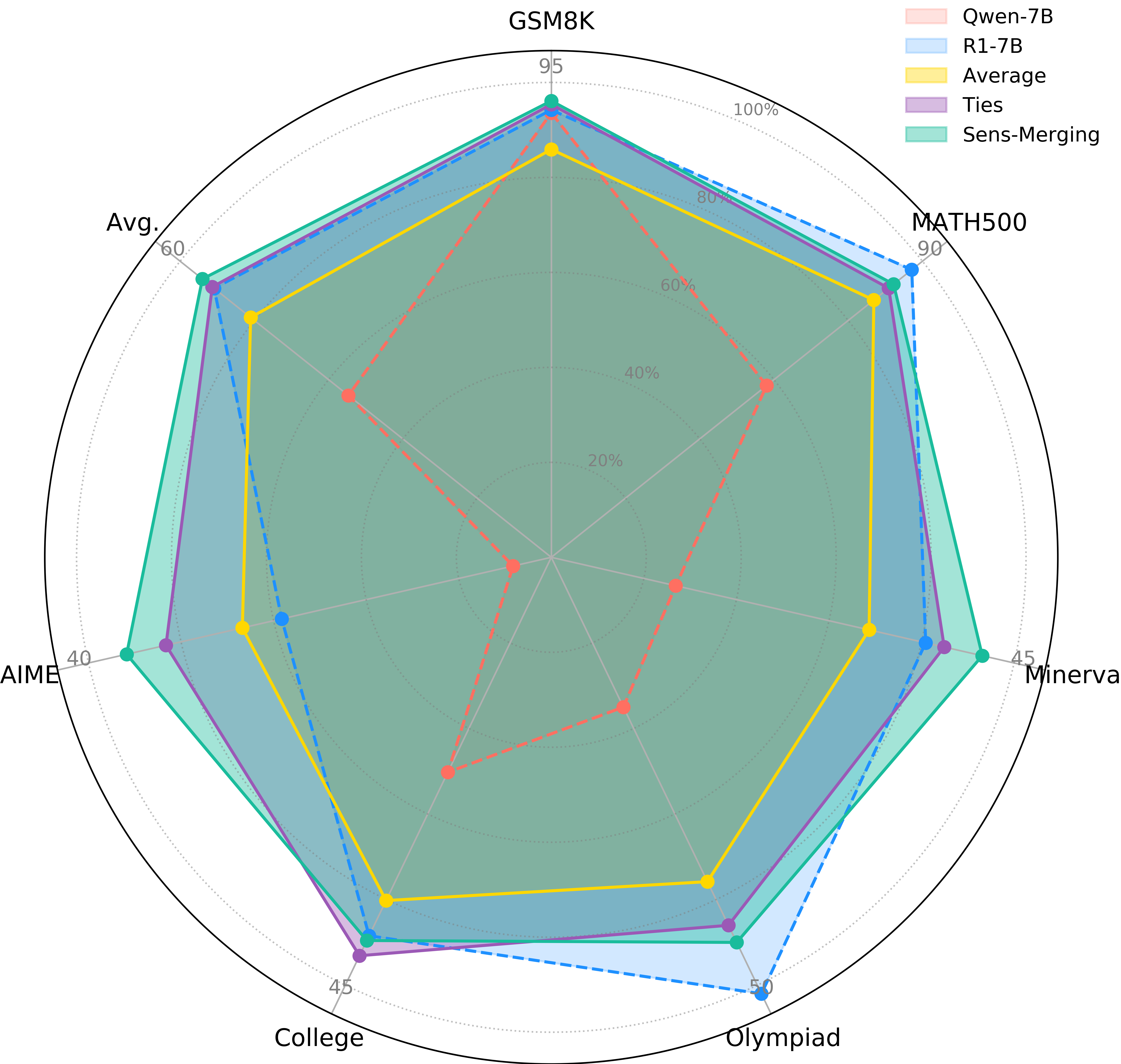}
    \end{subfigure}
    \caption{Performance of various model merging methods on DeepSeek-R1-7B and Qwen2.5-Math-7B models.}
    \label{fig:enter-label}
\end{figure}

{
\let\thefootnote\relax\footnotetext{Work in progress.}
\let\thefootnote\relax\footnotetext{* indicates equal contribution.}
}

\textbf{(TL;DR) Summary of our findings:}
\begin{itemize}
    \item Model merging is a highly efficient approach for long-to-short reasoning, as it directly operates on model parameters \textbf{without requiring additional training}.
    \item Task-vector based merging methods, especially like TA and TIES-Merging, can achieve long-to-short reasoning with around \textbf{50\% length reduction} alongside \textbf{accuracy parity or even marginal gains} on 7B models.
    \item SVD-based merging methods exhibit limited effectiveness, delivering moderate performance and serving as viable alternatives only when task vectors inherently possess low-rank spectral characteristics.
    \item Activation-based merging is the future, as it demonstrates impressive performance in terms of both reasoning accuracy (+1.9) and response length compression ratios (-49.8\%).
    \item Model merging methods applied to 1.5B-scale models remain effective on simple tasks. Smaller models struggle to learn long CoT reasoning ability through model merging.
    \item The merging of large-scale models (14B and 32B) poses significant challenges in simultaneously maintaining reasoning performance while substantially reducing response length.
\end{itemize}

\clearpage

\begin{abstract}
The transition from System 1 to System 2 reasoning in large language models (LLMs) has marked significant advancements in handling complex tasks through deliberate, iterative thinking. However, this progress often comes at the cost of efficiency, as models tend to overthink, generating redundant reasoning steps without proportional improvements in output quality. Long-to-Short (L2S) reasoning has emerged as a promising solution to this challenge, aiming to balance reasoning depth with practical efficiency. While existing approaches, such as supervised fine-tuning (SFT), reinforcement learning (RL), and prompt engineering, have shown potential, they are either computationally expensive or unstable. Model merging, on the other hand, offers a cost-effective and robust alternative by integrating the quick-thinking capabilities of System 1 models with the methodical reasoning of System 2 models.
In this work, we present a comprehensive empirical study on model merging for L2S reasoning, exploring diverse methodologies, including task-vector-based, SVD-based, and activation-informed merging. Our experiments reveal that model merging can reduce average response length by up to 55\% while preserving or even improving baseline performance. We also identify a strong correlation between model scale and merging efficacy with extensive evaluations on 1.5B/7B/14B/32B models. Furthermore, we investigate the merged model's ability to self-critique and self-correct, as well as its adaptive response length based on task complexity. Our findings highlight model merging as a highly efficient and effective paradigm for L2S reasoning, offering a practical solution to the overthinking problem while maintaining the robustness of System 2 reasoning.
This work can be found on Github \url{https://github.com/hahahawu/Long-to-Short-via-Model-Merging}.
\end{abstract}

\section{Introduction}
The development of large language models (LLMs) has transitioned from System 1 to System 2 reasoning \citep{yu2024distilling21,li202512surveyreasoning}, marked by the advent of advanced reasoning models such as OpenAI's o1/o3 \citep{o1,o3}, QwQ \citep{qwq}, and DeepSeek-R1 \citep{dpsk-r1}. System 1 reasoning models, such as GPT-4o \citep{4o}, LLaMA-3 \citep{grattafiori2024llama3herdmodels}, and DeepSeek-V3 \citep{deepseekai2025deepseekv3technicalreport}, are characterized by their capacity to generate straightforward, intuitive responses through rapid cognitive processing. However, this efficiency often limits their ability to handle intricate or multi-faceted tasks. In contrast, System 2 reasoning models prioritize deliberate, methodical analysis, engaging in iterative self-critique, error correction, and exhaustive evaluation prior to finalizing outputs. This systematic approach enhances their robustness, enabling superior performance on complex reasoning tasks.

Although significant advancements have been achieved in reasoning models, their practical efficiency remains limited by iterative thinking processes that often involve redundant reasoning steps and repetitive iterations. This inefficiency, commonly also referred to as the overthinking problem \citep{chen2025think23overthinkingo1like}, occurs when models excessively extend their deliberation without yielding proportional improvements in output quality. To mitigate this challenge, Long-to-Short (L2S) reasoning has garnered significant attention in recent research \citep{ma2025cotvalvelengthcompressiblechainofthoughttuning,xia2025tokenskipcontrollablechainofthoughtcompression}. Existing studies have explored diverse methods, including SFT \citep{xia2025tokenskipcontrollablechainofthoughtcompression}, reinforcement learning \citep{chen2025think23overthinkingo1like,luo2025o1prunerlengthharmonizingfinetuningo1like}, token-budget-aware inference \citep{han2025tokenbudgetawarellmreasoning} , prompt engineering \citep{luo2025o1prunerlengthharmonizingfinetuningo1like}, and model merging \citep{kimiteam2025kimik15scalingreinforcement}.
However, training-based methods are generally expensive, as they typically require the collection of aligned short and long answers for training. Prompting-based methods, on the other hand, have been shown to be unstable, as their performance is highly sensitive to changes in models and prompts. Research on model merging for long-to-short reasoning remains limited, with the only effort just focusing on the simplest average merging approach, which shows inferior performance \citep{kimiteam2025kimik15scalingreinforcement}.
Intuitively, if we view quick thinking and slow thinking as two distinct abilities developed through training on different tasks, model merging has been shown to effectively integrate these multi-task capabilities into a unified model \citep{task-arithmetic,DBLP:conf/nips/YadavTCRB23,DBLP:conf/icml/Yu0Y0L24}. Therefore, we argue that the merging approach represents a promising solution to the long-to-short reasoning problem.

In this work, we conduct a systematic empirical analysis of diverse model merging methodologies for long-to-short reasoning tasks. For example, we directly merge the quick-thinking System 1 model (like Qwen2.5-Math-7B) with its System 2 counterpart (like DeepSeek-R1-Distill-Qwen-7B) using existing merging approaches. Remarkably, our findings reveal that model merging serves as a highly efficient and effective paradigm for long-to-short optimization: it can reduce average response length up to 53\% while preserving baseline performance metrics, with certain configurations even achieving consistent performance improvements. Through a comprehensive analysis of various model merging methods on 7B models, we present several insightful observations: 1) task-vector-based merging methods demonstrate moderate effectiveness with minimal effort; 2) SVD-based merging methods generally yield unsatisfactory results; and 3) activation-informed merging shows promising performance.
Furthermore, our analysis reveals a strong correlation between model scale and merging efficacy. Specifically, smaller models (e.g., Qwen2.5-Math-1.5B) struggle to obtain the long-CoT reasoning ability through parameter fusion, whereas larger-scale models (e.g., Qwen2.5-14B/32B) can largely preserve reasoning performance, though the reduction in response length remains relatively insignificant.
More interestingly, we further investigate whether the merged model retains the ability for self-critique and self-correction, as well as how its response length correlates with the difficulty of the questions. The results reveal that the merged model essentially functions as a more intelligent slow-thinking model, capable of adjusting its output length based on the complexity of the problem and avoiding redundant reflections.

\section{Related Work}

\subsection{Long-to-Short Reasoning}
The evolution of large language models (LLMs) has transitioned from System 1 to System 2 reasoning \citep{li202512surveyreasoning}. System 1 models excel at generating quick, intuitive responses but often struggle with complex tasks. In contrast, System 2 models emphasize deliberate and methodical analysis, incorporating self-critique and error correction to enhance their performance on intricate reasoning tasks. However, their iterative processes can lead to inefficiencies, commonly referred to as the ``overthinking problem'' \citep{chen2025think23overthinkingo1like, yang2025thinkingoptimalscalingtesttimecompute}, where excessive deliberation fails to significantly improve the quality of the output. To mitigate this challenge, Long-to-Short (L2S) \citep{kimiteam2025kimik15scalingreinforcement} reasoning has garnered
significant attention in recent research.


An effective strategy for reducing inference costs without compromising accuracy involves compressing Chain of Thoughts (CoTs). TokenSkip \citep{xia2025tokenskipcontrollablechainofthoughtcompression} prioritizes tokens by semantic significance then conduct compression through selective omission. CoT-Valve \citep{ma2025cotvalvelengthcompressiblechainofthoughttuning} identifies a parameter space direction and adjusting step sizes systematically. Furthermore, to optimize computational resource allocation—prioritizing challenging tasks over simpler ones—O1-Pruner \citep{luo2025o1prunerlengthharmonizingfinetuningo1like} integrates sampling and reinforcement learning (RL) fine-tuning, while DAST \citep{shen2025dastdifficultyadaptiveslowthinkinglarge} introduces a difficulty metric based on token length budget to calibrate reward scores. Besides, there is a growing interest in leveraging target budgets to facilitate length control in conjunction with RL \citep{aggarwal2025l1controllinglongreasoning}. 
Additionally, prompt-based approaches have been extensively explored in recent works\citep{kimiteam2025kimik15scalingreinforcement, luo2025o1prunerlengthharmonizingfinetuningo1like}. Whether through short responses in few-shot settings or encouraging the model to provide ``concise'’ replies in demonstrations \citep{munkhbat2025selftrainingelicitsconcisereasoning}, these attempts have demonstrated instability, as their performance is highly sensitive to variations in models and prompts.

\subsection{Model Merging}
Model merging \citep{merge_survey}, as a complementary approach to training-based methods, enables the integration of multiple task-specialized models into a unified model \citep{model_soup,DBLP:conf/nips/YadavTCRB23}. This approach enhances model performance on individual tasks by merging checkpoints without the need for additional training \citep{DBLP:conf/nips/YadavTCRB23, task-arithmetic}, mitigates the issue of catastrophic forgetting \citep{mitigating_cf}, and achieves the long-to-short reasoning \citep{kimiteam2025kimik15scalingreinforcement}.
In this work, we categorize existing model merging approaches into three groups: 1) vanilla task-vector-based merging methods \citep{model_soup,task-arithmetic,DBLP:conf/nips/YadavTCRB23,DBLP:conf/icml/Yu0Y0L24}, which represent fine-tuned features as task vectors and perform arithmetic operations on them to derive the merged model; 2) SVD-based merging methods \citep{DBLP:conf/nips/LuF0QC024,liu2025loremergingexploringlowrankestimation}, characterized by their ability to identify and exploit the low-rank features of task vectors; and 3) Activation-based methods \citep{nobari2025activationinformedmerginglargelanguage,liu2025sensmergingsensitivityguidedparameterbalancing}, which utilize input activations to assign varying importance scores to the merging models.

Besides, MoE-based merging methods \citep{sukhbaatar2024branchtrainmixmixingexpertllms,liu20251bitmergingdynamicquantizedmerging} also play an important role in this field. However, we exclude the discussion of these methods as they typically involve modifications to the model architecture and model size.
\section{Background}
\subsection{Model Merging}
Model merging seeks to integrate multiple fine-tuned (FT) models, derived from a pre-trained (PT) model $\theta_0$, into a unified model that consolidates knowledge from diverse sources. Given $K$ FT models to be merged, denoted as $\theta_1, \dots, \theta_K$, the goal is to produce a single model $\theta_M$ that inherits the capabilities of the individual models.

\paragraph{Average Merging}
Average merging \citep{model_soup} is a simple and effective method to enhance overall performance by  performing an arithmetic average of the model weights. It reduces variance by smoothing random errors, especially when base models are diverse and exhibit low bias. However, its effectiveness depends on the quality and diversity of the base models; high bias across models limits its improvement potential.

\paragraph{Task Arithmetic (TA)}
In most existing task-vector-based approaches, the base model $\theta_0$ is essential for computing task vectors \citep{task-arithmetic}, which generally encapsulate the knowledge acquired during fine-tuning. A task vector is defined as the parameter shift between an FT model and its corresponding base model, expressed as $\delta_k = \theta_k - \theta_0$. The merged model $\theta_M$ is then obtained by aggregating the task vectors into the base model, as $\theta_M = \theta_0 + \sum_{k=1}^K \lambda_k \cdot \delta_k$, where $\lambda_k$ represents the weight coefficient, which can either be manually set as a constant or determined through optimization.

\paragraph{TIES-Merging}
TIES-Merging \citep{DBLP:conf/nips/YadavTCRB23} is an efficient method for integrating parameters from multiple FT models, addressing redundancy and conflicts. Its key steps include: (1) pruning parameters, retaining significant deviations from pre-trained weights; (2) resolving conflicts via majority voting or alignment; and (3) weighted aggregation of significant parameters to form the final model. This approach reduces noise and enhances generalization, particularly for integrating fine-tuned large language models (LLMs) across related tasks.

\paragraph{DARE} \citep{DBLP:conf/icml/Yu0Y0L24} DARE Merging is a lightweight approach, whose core steps include: (1) randomly dropping redundant parameters (e.g., those with minimal gradient changes) to reduce noise; (2) adjusting the direction of retained parameters to resolve conflicts between models; and (3) performing weighted integration of key parameters to preserve essential knowledge. This method significantly reduces computational overhead while maintaining model performance, making it suitable for rapid fusion of multi-task fine-tuned models.


\paragraph{AIM}
The core idea of AIM \citep{nobari2025activationinformedmerginglargelanguage} is to utilize activation space information to identify and protect critical weights in PT models, thereby minimizing alterations to these weights during the merging process. This approach is akin to weight regularization techniques in Continual Learning (CL), aiming to prevent catastrophic forgetting, which is the loss of performance on old tasks when learning new ones. The effectiveness of AIM across various merging methods and benchmarks underscores the importance of activation space information in model merging. As a supplementary method, AIM can be integrated with existing merging techniques to significantly enhance the performance and robustness of merged models.

\paragraph{LoRE-Merging} Existing merging methods rely on sparsely estimated task vectors but face two key limitations: dependence on base model parameters and task vector interference. LoRE-Merging \citep{liu2025loremergingexploringlowrankestimation}, a low-rank estimation framework, is proposed to address such issues. It constructs an approximate base model and low-rank task vectors via optimization, minimizing discrepancies between fine-tuned and merged models. Using coordinate descent and singular value thresholding, LoRE-Merging reduces task vector interference, demonstrating the effectiveness of low-rank estimation in model merging.

\paragraph{Twin-Merging} Performance gaps between merged and fine-tuned models stem from conflicts among models and diverse testing data. Twin-Merging \citep{DBLP:conf/nips/LuF0QC024} resolves this by categorizing expert knowledge into generalizable shared knowledge and task-specific knowledge. Through compression and difference extraction, this knowledge is modularized. A router then dynamically integrates shared and task-specific knowledge based on input, similar to the Mixture of Experts approach, allowing for flexible adjustments. In our study, we eliminate the router training and directly utilize its singular value decomposition (SVD) merging part.


\paragraph{Sens-Merging}
Sens-Merging \citep{liu2025sensmergingsensitivityguidedparameterbalancing} focuses on the varying importance of parameters within and across tasks during model merging. It operates at two levels: (1) within individual tasks, where parameter sensitivity analysis identifies critical layers impacting performance, and (2) across tasks, where task sensitivity analysis prioritizes models that enhance others' performance. By combining these analyses, Sens-Merging derives merging coefficients for fine-grained parameter control, enabling effective layer-wise merging. It also serves as a plug-and-play enhancement to task vector-based merging, improving flexibility and performance.


\subsection{LLM Reasoning}
We divide the CoT-based LLM reasoning into two categories, including short-CoT (quick-thinking) reasoning and long-CoT (slow-thinking) reasoning. Given an input $X$, the quick-thinking reasoning models directly output the final answer, represented as $\{\theta, X\} \rightarrow A$. In contrast, slow-thinking models generate both the extensive thinking process and the final answer, denoted as $\{\theta, X\} \rightarrow [T,A]$. While the reasoning process in slow-thinking models often enhances the quality of the final answer, the lengthy context significantly impairs inference efficiency, particularly when the reasoning content contains substantial repetitions or unnecessary reflections.

\section{Exploring Model Merging Methods for Long-to-Short Reasoning}
In this section, we explore the effect of model merging on the most frequently used 7B reasoning models, i.e. Qwen2.5-Math-7B and DeepSeek-R1-Distill-Qwen-7B \citep{dpsk-r1}. Then, we reveal the effectiveness of model merging on achieving long-to-short reasoning.

\subsection{Experiment Setup}
We conduct evaluations on popular reasoning datasets: GSM8K, MATH500, Minerva Math, Olympiadbench, College Math and AIME24. We utilize the public evaluation toolkit\footnote{https://github.com/QwenLM/Qwen2.5-Math} provided by Qwen for better reproducibility.
The quick-thinking models are evaluated with few shots and the slow-thinking models are in a zero-shot setting, as recommended in \citet{dpsk-r1}.
For activation-based merging methods, we adopt the \textbf{s1K} \citep{s1} dataset which contains the high-quality aligned quick-thinking answer and slow-thinking answer for each question. The maximum length for quick-thinking models and slow-thinking models are 8K and 10K, respectively. We also establish a training-based baseline using DPO, where DeepSeek-R1-7B is trained on the s1K dataset with the short answers as positive samples.

\begin{table}[t!]
\fontsize{9}{10} \selectfont
    \centering
    \def\arraystretch{1,2}
    \begin{tabular}{cccccccc}
    \toprule[0.8pt]
       \diagbox{Method}{Bench} & GSM8K & MATH500 & \makecell[c]{Minerva\\Math} & \makecell[c]{Olympiad\\Bench} & \makecell[c]{College\\Math} & AIME24 & \multirow{1}{*}{Avg.} \\
       \hline
       \multirow{2}{*}{Qwen2.5-Math-7B} & 88.9 & 52.2 & 12.1 & 17.5 & 22.6 & 3.3 & \multirow{2}{*}{32.8} \\
       & \cellcolor{brown!10} (130.2) & \cellcolor{brown!10}(526.2) & \cellcolor{brown!10}(956.7) & \cellcolor{brown!10}(1259.2) & \cellcolor{brown!10}(794.8) & \cellcolor{brown!10} (1528.0) \\
       \multirow{2}{*}{DeepSeek-R1-7B} & 89.3 & 87.4 & 36.4 & 51.0 & 39.8 & 23.3 & \multirow{2}{*}{54.5} \\
       & \cellcolor{brown!10}(1062.0) & \cellcolor{brown!10}(2825.9) & \cellcolor{brown!10}(3055.9) & \cellcolor{brown!10}(5793.8) & \cellcolor{brown!10}(2461.6) & \cellcolor{brown!10}(8675.4) \\
       \cdashline{1-8}
       \multirow{2}{*}{DPO} & 89.6 & 88.4 & 33.8 & 48.6 & 39.7 & 40.0 & 56.7\\
       & \cellcolor{brown!10}(983.6) & \cellcolor{brown!10} (2587.1) & \cellcolor{brown!10} (2304.9) & \cellcolor{brown!10} (4932.3) & \cellcolor{brown!10} (2073.5) & \cellcolor{brown!10} (7029.5) & \cellcolor{brown!10} (-15.0\%)\\
       \rowcolor{lightgray}
       \multicolumn{8}{c}{\textit{Task-vector based Merging Methods}} \\
       \multirow{2}{*}{Average Merging} & 81.6 & 78.2 & 30.9 & 37.9 & 36.1 & 26.7 & \cellcolor{green!10}48.6 \\
       & \cellcolor{brown!10} (636.4) & \cellcolor{brown!10} (1416.7) & \cellcolor{brown!10} (2277.4) & \cellcolor{brown!10} (3202.2) & \cellcolor{brown!10} (2065.7) & \cellcolor{brown!10} (5964.8) & \cellcolor{brown!10} (-34.6\%)\\
       \multirow{2}{*}{Task Arithmetic} & 90.5 & 83.4 & 41.9 & 45.0 & 40.3 & 20.0 & \cellcolor{green!10}53.5 \\
       & \cellcolor{brown!10} (617.5) & \cellcolor{brown!10} (1617.4) & \cellcolor{brown!10} (1416.2) & \cellcolor{brown!10} (2650.6) & \cellcolor{brown!10} (1443.9) & \cellcolor{brown!10} (3594.6) & \cellcolor{brown!10} (-48.7\%) \\
       \multirow{2}{*}{TIES-Merging} & 90.6 & 81.8 & 38.2 & 43.0 & 41.9 & 33.3 & \cellcolor{green!10}54.8\\
       & \cellcolor{brown!10} (552.2) & \cellcolor{brown!10} (1492.9) & \cellcolor{brown!10} (1349.2) & \cellcolor{brown!10} (2473.7) & \cellcolor{brown!10} (1287.8) & \cellcolor{brown!10} (3302.1) & \cellcolor{brown!10} (-53.0\%)\\
       \multirow{2}{*}{DARE} & 84.0 & 75.4 & 29.4 & 35.7 & 36.2 & 23.3 & \cellcolor{green!10}47.3\\
       & \cellcolor{brown!10} (815.6) & \cellcolor{brown!10} (2237.1) & \cellcolor{brown!10} (2317.8) & \cellcolor{brown!10} (3266.3) & \cellcolor{brown!10} (2072.1) & \cellcolor{brown!10} (3803.1) & \cellcolor{brown!10} (-30.6\%)\\
       \multirow{2}{*}{DARE-TA} & 87.9 & 84.0 & 29.4 & 41.6 & 38.3 & 26.7 & \cellcolor{green!10}51.3 \\
       & \cellcolor{brown!10} (600.4) & \cellcolor{brown!10} (1703.9) & \cellcolor{brown!10} (1567.2) & \cellcolor{brown!10} (2774.8) & \cellcolor{brown!10} (1533.9) & \cellcolor{brown!10} (3454.2) & \cellcolor{brown!10} (-46.0\%)\\
       \multirow{2}{*}{DARE-TIES} & 89.6 & 82.4 & 37.5 & 41.3 & 40.2 & 23.3 & \cellcolor{green!10}52.4 \\
       & \cellcolor{brown!10} (584.3) & \cellcolor{brown!10} (1589.4) & \cellcolor{brown!10} (1307.4) & \cellcolor{brown!10} (2655.5) & \cellcolor{brown!10} (1378.5) & \cellcolor{brown!10} (3579.9) & \cellcolor{brown!10} (-50.4\%)\\
       \hline
       \rowcolor{lightgray}
       \multicolumn{8}{c}{\textit{SVD-based Merging Methods}} \\
       \multirow{2}{*}{LoRE-Merging} & 84.8 & 80.8 & 37.5 & 40.9 & 35.5 & 30 & \cellcolor{green!10}51.6\\
       & \cellcolor{brown!10} (531.7) & \cellcolor{brown!10} (1819.1) & \cellcolor{brown!10} (1944.9) & \cellcolor{brown!10} (3072.3) & \cellcolor{brown!10} (1875.9) & \cellcolor{brown!10} (3691.6) & \cellcolor{brown!10} (-41.6\%)\\
       \multirow{2}{*}{Twin-Merging} & 87.6 & 79.6 & 31.2 & 40.0 & 38.1 & 26.7 & \cellcolor{green!10}50.5 \\
       & \cellcolor{brown!10} (778.1) & \cellcolor{brown!10} (1997.3) & \cellcolor{brown!10} (2079.9) & \cellcolor{brown!10} (3098.3) & \cellcolor{brown!10} (1870.5) & \cellcolor{brown!10} (3768.9) & \cellcolor{brown!10} (-35.8\%)\\
       \rowcolor{lightgray}
       \multicolumn{8}{c}{\textit{Activation-based Merging Methods}} \\
       \multirow{2}{*}{AIM-TIES} & 90.8 & 83.0 & 40.8 & 46.4 & 42.3 & 26.7 & \cellcolor{green!10}55.0 \\
       & \cellcolor{brown!10} (540.6) & \cellcolor{brown!10} (1374.5) & \cellcolor{brown!10} (1229.8) & \cellcolor{brown!10} (2323.8) & \cellcolor{brown!10} (1249.6) & \cellcolor{brown!10} (3265.9) & \cellcolor{brown!10} (-55.3\%)\\
       \multirow{2}{*}{Sens-Merging} &  91.3 & 83.0 & 41.9 & 45.0 & 40.3 & 36.7 & \cellcolor{green!10} 56.4 \\
       & \cellcolor{brown!10} (600.7) & \cellcolor{brown!10} (1650.2) & \cellcolor{brown!10} (1345.9) & \cellcolor{brown!10} (2673.2) & \cellcolor{brown!10} (1446.4) & \cellcolor{brown!10} (3245.9) & \cellcolor{brown!10} (-49.8\%)\\
    \bottomrule[0.8pt]
    \end{tabular}
    \caption{Evaluations of different model merging methods on Qwen-7B models. The number in \colorbox{brown!10}{()} indicates the average response length on the dataset.}
    \label{tab:7b_merge}
\end{table}

\subsection{Task-Vector based Merging Works}
We first explore the typical task-vector based merging methods: \textbf{Average Merging}, \textbf{Task Arithmetic}, \textbf{TIES-Merging} and \textbf{DARE}. We consider the average merging as a special case of task-arithmetic merging.

As shown in the second part of Table \ref{tab:7b_merge}, task-vector based methods demonstrates significant efficacy in compressing output length while preserving reasoning accuracy. The baseline of average merging reduces response length by 34.6\% relative to the reasoning model while improving accuracy by 15.8\% over the quick-thinking baseline. Although this approach incurs marginal performance degradation compared to the pure reasoning model, advanced merging paradigms effectively mitigate this tradeoff. Notably, TA and TIES-Merging achieve comprehensive optimization, delivering 48-53\% length reduction alongside accuracy parity (within 1.0\%) or even marginal gains (+0.3\% on the average score). This finding is particularly significant as it demonstrates that long-to-short reasoning can be accomplished with minimal computational effort.
Furthermore, DARE, as a plug-and-play method, can be combined with other task-vector-based merging approaches. However, it consistently underperforms compared to the original merging methods. We hypothesize two potential reasons for this: 1) as noted by \citet{DBLP:conf/icml/Yu0Y0L24}, the performance of DARE is constrained by parameter shifts being within 0.002, whereas the parameter shifts between Qwen-7B models significantly exceed this threshold; 2) the random dropping of certain task vectors, a key feature of DARE, might inadvertently remove critical parameters. Supporting this assumption, we empirically observe that the drop ratio in DARE should be less than 0.5, rather than the larger value of 0.9 recommended by \citet{DBLP:conf/icml/Yu0Y0L24}.

Upon closely examining the performance scores across each dataset, we present the following observations: 1) Accuracy improvements over the reasoning model are more readily achieved on datasets where the quick-thinking model and the reasoning model exhibit similar performance levels, such as GSM8K and College Math; 2) The merged model demonstrates limited capacity to surpass the reasoning model when substantial initial performance disparities exist between the base models, such as MATH500 and OlympiadBench; 3) The reduction in response length is more pronounced on complex datasets, such as AIME24 and OlympiadBench.

\begin{mybox}[colback=gray!10]{Takeaway 1}
\textbf{\textit{Task-vector based merging methods, especially like TA and TIES-Merging, can achieve long-to-short reasoning with around 50\% length reduction alongside accuracy parity or even marginal gains.}}
\end{mybox}

\subsection{SVD-based Merging Underperforms}
Based on the observation that task vectors often exhibit a limited number of dominant singular values, recent SVD-based merging methods aim to address task vector interference through low-rank approximation. In our investigation, we examined the following SVD-based merging methods: \textbf{LoRE-Merging} and \textbf{Twin-Merging}.

As shown in the third section of Table \ref{tab:7b_merge}, both LoRE-Merging and Twin-Merging outperform the average merging method in terms of both length compression and reasoning accuracy, highlighting their effectiveness in long-to-short reasoning tasks. However, their performance falls short compared to advanced task-vector-based methods, such as TA and TIES-Merging. We hypothesize that the effectiveness of these approaches is highly dependent on the distribution of the domain singular values of the task vectors. Notably, we observe that the singular value distributions of the candidate base models deviate from the ideal distribution. One more interesting finding is that, although the overall performance of SVD-based merging methods is moderate, they appear to consistently perform well on complex tasks, such as AIME24.

\begin{mybox}[colback=gray!10]{Takeaway 2}
\textbf{\textit{SVD-based merging methods exhibit limited effectiveness, delivering moderate performance and serving as viable alternatives only when task vectors inherently possess low-rank spectral characteristics.}}
\end{mybox}

\subsection{Activation-based Merging is the Future}
Inspired by prior research in model quantization \citep{awq} and pruning \citep{wanda,liu2025optishearefficientadaptivepruning}, activation-based merging methods have also been extensively studied.
As observed in task-based merging methods \citep{task-arithmetic,nobari2025activationinformedmerginglargelanguage}, the choice of coefficients plays a crucial role in achieving effective merging. The most intuitive approach is to leverage activations to search for optimal coefficients.
In this work, we evaluate two recent approaches: \textbf{AIM} and \textbf{Sens-Merging}. For our experiments, we utilize the s1K dataset as the calibration dataset, as it provides high-quality, well-aligned short and long responses for each question.


For Sens-Merging, we use quick-thinking answers as calibration data for the R1 model to enhance its ability to generate concise responses and reduce reasoning path length. Meanwhile, slow-thinking answers are selected as calibration data for the Qwen Math model to promote its deep reasoning capabilities. We randomly choose 100 samples from the s1K dataset for calibration, making the sensitivity score computation more efficient.
Similarly, we also use the sampled data as the calibration data of AIM-TIES which operates on the merged model from TIES-Merging and Qwen base model.

As evidenced in Table \ref{tab:7b_merge}, both activation-based methodologies surpass the DeepSeek-R1-7B baseline across comprehensive metrics, achieving superior average scores paired with significant response length reduction. Specifically, applying AIM to the merged model from TIES-Merging yields further enhancements - elevating performance by 0.2 points while raising the compression ratio to 55.3\%. For Sens-Merging, it even achieves the comparable reasoning performance with DPO while significantly reducing the response length by approximately 50\%.

However, there are certain limitations in existing activation-based methods. For instance, Sens-Merging requires gradient computation during the backward pass, which can impact its efficiency. In contrast, AIM utilizes activations solely during the forward pass, enhancing the efficiency of activation usage. Nonetheless, the performance of activation-based methods is inherently reliant on the choice of calibration data.
In our experiments, we also evaluated AIM and Sens-Merging with alternative calibration datasets but observed inferior performance. While existing activation-based methods have achieved notable success, further research into more robust and efficient activation-based merging techniques remains a critical direction for future work.

\begin{mybox}[colback=gray!10]{Takeaway 3}
\textbf{\textit{Activation-based merging methods demonstrate superior performance in terms of both reasoning accuracy and response length compression rates; however, their effectiveness is certainly dependent on the choice of the calibration dataset.}}
\end{mybox}

\section{Analysis on Models with Different Scales}
Apart from the evaluations on Qwen-7B models, we further investigate the effectiveness of model merging across different model scales, including the smaller Qwen-1.5B model and larger models such as Qwen-14B and Qwen-32B. To simplify the evaluation process, we only test on GSM8K, MATH500 and AIME24 datasets in this section.

\begin{table}[t!]
    \centering
    \def\arraystretch{1,2}
    \begin{tabular}{cccccccc}
    \toprule[0.8pt]
    \multirow{2}{*}{\diagbox{Method}{Bench}} & \multicolumn{3}{c}{Accuracy} & & \multicolumn{3}{c}{Length \& Reflections}  \\
    \cline{2-4} \cline{6-8}
    & GSM8K & MATH500 & AIME24 & & GSM8K & MATH500 & AIME24 \\
    \hline
    \multirow{2}{*}{Qwen2.5-Math-1.5B} & \multirow{2}{*}{75.9} & \multirow{2}{*}{36.2} & \multirow{2}{*}{0.0} & & 118.1 & 411.0 & 864.9\\
    & & & & & \cellcolor{brown!10} [0] &\cellcolor{brown!10} [0] &\cellcolor{brown!10} [0] \\
    \multirow{2}{*}{DeepSeek-R1-1.5B} & \multirow{2}{*}{76.8} & \multirow{2}{*}{69.6} & \multirow{2}{*}{20.0} & & 2744.3 & 4510.0 & 8952.3 \\
    & & & & & \cellcolor{brown!10} [1315] & \cellcolor{brown!10} [499] & \cellcolor{brown!10} [30] \\
    \hline
    \multirow{2}{*}{Average Merging} & \multirow{2}{*}{26.6} & \multirow{2}{*}{12.2} & \multirow{2}{*}{0.0} & & 3428.7 & 3890.6 & 3977.3\\
    & & & & & \cellcolor{brown!10} [774] & \cellcolor{brown!10} [306] & \cellcolor{brown!10} [16] \\
    Task Arithmetic & 74.5 & 62.6 & 10.0 & & 549.7 & 1619.4 & 3395.3 \\
    & & & & & \cellcolor{brown!10} [24] &\cellcolor{brown!10} [51] &\cellcolor{brown!10} [9] \\
    TIES-Merging & 75.9 & 64.2 & 10.0 & & 393.8 & 1102.4 & 2818.7\\
    & & & & & \cellcolor{brown!10} [24] &\cellcolor{brown!10} [53] &\cellcolor{brown!10} [3] \\
    LoRE-Merging & 64.1 & 50.8 & 6.7 & & 763.8 & 2248.4 & 3833.3 \\
    & & & & & \cellcolor{brown!10} [118] &\cellcolor{brown!10} [197] &\cellcolor{brown!10} [15] \\
    Sens-Merging & 81.3 & 69.8 & 16.7 & & 340.3 & 852.8 & 2102.6 \\
    & & & & & \cellcolor{brown!10} [9] &\cellcolor{brown!10} [8] &\cellcolor{brown!10} [0] \\
    \bottomrule[0.8pt]
    \end{tabular}
    \caption{Evaluations of various model merging methods on Qwen-1.5B models. The number in \colorbox{brown!10}{[]} indicates the number of reflective responses on the dataset.}
    \label{tab:1.5B_res}
\end{table}

\subsection{Smaller models struggle to learn from model merging}
The evaluation results are presented in Table \ref{tab:1.5B_res}. Firstly, we find the task-vector-based methods, such as TA and TIES-Merging, remain effective in reducing the response length while maintaining reasoning performance. The activation-based method, i.e. Sens-Merging, consistently achieves the best performance across the board. But unfortunately, the overall performance of model merging on 1.5B models lags behind that observed on 7B-scale models, particularly on more complex tasks. For instance, on the AIME24 dataset, consistent performance drops are observed across all merging methods. Besides, we unexpectedly find that the number of reflective responses\footnote{The definition of reflective responses is provided in Section \ref{sec:analysis}.} is even negatively correlated with the final performance. This suggests that the reflection actions learned by the merged models are ``false reflections'', which ultimately lead to incorrect final answers.
Consequently, we argue that smaller 1.5B models face significant challenges in acquiring long CoT reasoning abilities from stronger reasoner models through model merging. This observation aligns with the findings reported in \citet{li2025smallmodelsstrugglelearn}.

\begin{mybox}[colback=gray!10]{Takeaway 4}
\textbf{\textit{Model merging methods applied to 1.5B-scale models, such as TA, TIES-Merging and Sens-Merging, remain effective on simple tasks. Smaller models struggle to learn long CoT reasoning ability through model merging.}}
\end{mybox}

\begin{table}[t]
   \fontsize{8.5}{11} \selectfont
    \centering
    \def\arraystretch{1,2}
    \begin{tabular}{ccccccccc}
    \toprule[0.8pt]
    \multirow{2}{*}{\diagbox{Method}{Bench}} & \multicolumn{3}{c}{Accuracy} & & \multicolumn{4}{c}{Length \& Reflections}  \\
    \cline{2-4} \cline{6-9}
    & GSM8K & MATH500 & AIME24 & & GSM8K & MATH500 & AIME24 & Avg. \\
    \hline
    \multirow{2}{*}{Qwen2.5-14B} & \multirow{2}{*}{90.8} & \multirow{2}{*}{47.2} & \multirow{2}{*}{3.3} & & 138.5 & 522.9 & 1493.8 & \multirow{2}{*}{-77.8\%}\\
    & & & & & \cellcolor{brown!10} [2;1.0] &\cellcolor{brown!10} [0;0.0] &\cellcolor{brown!10} [0;0.0] \\
    \multirow{2}{*}{DeepSeek-R1-14B} & \multirow{2}{*}{91.9} & \multirow{2}{*}{89.0} & \multirow{2}{*}{50.0} & & 558.8 & 2314.1 & 7724.8 & \multirow{2}{*}{0} \\
    & & & & & \cellcolor{brown!10} [24;2.8] & \cellcolor{brown!10} [473;9.2] & \cellcolor{brown!10} [30;32.4] \\
    \hline
    \multirow{2}{*}{Average Merging} & \multirow{2}{*}{91.5} & \multirow{2}{*}{88.2} & \multirow{2}{*}{46.7} & & 642.8 & 2925.1 & 7922.7 & \multirow{2}{*}{+14.7\%}\\
    & & & & & \cellcolor{brown!10} [372;3.9] & \cellcolor{brown!10} [457;12.8] & \cellcolor{brown!10} [30;37.2] \\
    \multirow{2}{*}{Task Arithmetic} & \multirow{2}{*}{94.1} & \multirow{2}{*}{85.4} & \multirow{2}{*}{33.3} & & 565.9 & 2191.4 & 6873.6 & \multirow{2}{*}{-5.0\%}\\
    & & & & & \cellcolor{brown!10} [861;3.0] &\cellcolor{brown!10} [472;6.9] &\cellcolor{brown!10} [29;17.3] \\
    \multirow{2}{*}{TIES-Merging} & \multirow{2}{*}{93.4} & \multirow{2}{*}{84.6} & \multirow{2}{*}{26.7} & & 503.1 & 2169.7 & 6953.7 & \multirow{2}{*}{-8.7\%}\\
    & & & & & \cellcolor{brown!10} [991;2.4] &\cellcolor{brown!10} [455;6.9] &\cellcolor{brown!10} [29;23.6]\\
    \multirow{2}{*}{AIM-TIES} & \multirow{2}{*}{93.0} & \multirow{2}{*}{86.4} & \multirow{2}{*}{50.0} & & 534.4 & 2260.0 & 7129.0 & \multirow{2}{*}{-4.8\%}\\
    & & & & & \cellcolor{brown!10} [568;2.8] &\cellcolor{brown!10} [462;8.1] &\cellcolor{brown!10} [30;30.5] \\
    \multirow{2}{*}{Sens-Merging} & \multirow{2}{*}{94.2} & \multirow{2}{*}{89.4} & \multirow{2}{*}{53.3} & & 654.4 & 2601.5 & 7435.1 & \multirow{2}{*}{+8.6\%}\\
    & & & & & \cellcolor{brown!10} [714;3.1] &\cellcolor{brown!10} [483;8.8] &\cellcolor{brown!10} [29;29.9] \\
    \bottomrule[0.8pt]
    \end{tabular}
    \caption{Evaluations of various model merging methods on Qwen-14B models. The number $A$ in \colorbox{brown!10}{[A;B]} indicates the number of reflective responses on the dataset and number $B$ indicates the average frequency of reflection keywords appearing in each response.}
    \label{tab:14B_res}
\end{table}
\subsection{Length reduction is challenging on large-scale models}
We evaluate the performance of several advanced model merging methods on 14B and 32B models. Table \ref{tab:14B_res} reports the results for Qwen 14B models. While reasoning performance is largely preserved after model merging, the length reduction ratios are less significant compared to those observed in smaller-scale models. Notably, methods such as average merging and Sens-Merging even lead to an increase in response length, although Sens-Merging consistently enhances reasoning accuracy across all benchmarks. Similar trends are observed for 32B models, as shown in Table \ref{tab:32B_res}.
In the experiments with 14B and 32B models, we attempted to fine-tune the hyper-parameters of the merging methods to achieve a significant reduction in response length. For instance, a length reduction ratio of 58.6\% was achieved by setting the coefficient of TA merging to 0.3. However, this configuration resulted in a substantial performance drop (GSM8K: +1.2; MATH500: -7.2; AIME24: -30.0). Conversely, a notable observation is that the performance on GSM8K remaines consistently comparable or even marginally higher under most settings. We attribute this to the slight performance variation of the base models on GSM8K, which differs by no more than 3.5 points. Since both Qwen2.5-14B and Qwen2.5-32B are general-purpose models, unlike domain-specific models such as Qwen2.5-Math-1.5B/7B that are extensively trained on mathematical tasks, there exists a significant performance disparity between these general-purpose models and the R1-distilled models on complex mathematical tasks like MATH500 and AIME, with gaps of approximately 40\%. This substantial gap makes it challenging for the merged models to surpass the R1-distilled models, as they struggle to provide additional informative knowledge to the already superior models. This hypothesis is inline with the finding in model ensembling \citep{unite}.

\begin{table}[t!]
   \fontsize{8.5}{10} \selectfont
    \centering
    \def\arraystretch{1,2}
    \begin{tabular}{ccccccccc}
    \toprule[0.8pt]
    \multirow{2}{*}{\diagbox{Method}{Bench}} & \multicolumn{3}{c}{Accuracy} & & \multicolumn{4}{c}{Length \& Reflections}  \\
    \cline{2-4} \cline{6-9}
    & GSM8K & MATH500 & AIME24 & & GSM8K & MATH500 & AIME24 & Avg. \\
    \hline
    \multirow{2}{*}{Qwen2.5-32B} & \multirow{2}{*}{92.3} & \multirow{2}{*}{55.4} & \multirow{2}{*}{6.7} & & 130.3 & 497.2 & 1127.9 & \multirow{2}{*}{-83.2\%}\\
    & & & & & \cellcolor{brown!10} [3;0.0] &\cellcolor{brown!10} [0;0.0] &\cellcolor{brown!10} [0;0.0] \\
    \multirow{2}{*}{DeepSeek-R1-32B} & \multirow{2}{*}{95.7} & \multirow{2}{*}{89.2} & \multirow{2}{*}{60.0} & & 822.6 & 2621.8 & 7194.2 & \multirow{2}{*}{0}\\
    & & & & & \cellcolor{brown!10} [1287;3.3] & \cellcolor{brown!10} [497;10.2] & \cellcolor{brown!10} [30;34.3] \\
    \hline
    \multirow{2}{*}{Average Merging} & \multirow{2}{*}{93.6} & \multirow{2}{*}{89.2} & \multirow{2}{*}{66.7} & & 662.2 & 2656.4 & 6966.8 & \multirow{2}{*}{-7.1\%}\\
    & & & & & \cellcolor{brown!10} [570;3.8] & \cellcolor{brown!10} [489;10.5] & \cellcolor{brown!10} [30;27.1] \\
    \multirow{2}{*}{Task Arithmetic} & \multirow{2}{*}{95.5} & \multirow{2}{*}{87.6} & \multirow{2}{*}{50.0} & & 580.1 & 2168.7 & 6452.7 & \multirow{2}{*}{-19.0\%} \\
    & & & & & \cellcolor{brown!10} [991;2.4] &\cellcolor{brown!10} [455;6.9] &\cellcolor{brown!10} [29;23.6] \\
    \multirow{2}{*}{TIES-Merging} & \multirow{2}{*}{95.4} & \multirow{2}{*}{87.8} & \multirow{2}{*}{43.3} & & 516.9 & 1898.7 & 5832.3 & \multirow{2}{*}{-27.9\%}\\
    & & & & & \cellcolor{brown!10} [763;2.0] &\cellcolor{brown!10} [391;4.2] &\cellcolor{brown!10} [24;11.3] \\
    \bottomrule[0.8pt]
    \end{tabular}
    \caption{Evaluations of various model merging methods on Qwen-32B models.}
    \label{tab:32B_res}
\end{table}
\begin{table}[t!]
   \fontsize{9}{10} \selectfont
    \centering
    \def\arraystretch{1,2}
    \begin{tabular}{cccccccc}
    \toprule[0.8pt]
    {\diagbox{Method}{Bench}} & \multicolumn{3}{c}{Accuracy} & & \multicolumn{3}{c}{Length}  \\
    \cline{2-4} \cline{6-8}
    & GSM8K & MATH500 & AIME24 & & GSM8K & MATH500 & AIME24 \\
    \hline
    {Qwen2.5-32B} & {92.3} & {55.4} & {6.7} & & 130.3 & 497.2 & 1127.9 \\
    {DeepSeek-R1-32B} & {95.7} & {89.2} & {60.0} & & 822.6 & 2621.8 & 7194.2 \\
    QwQ-32B & 96.1 & 90.8 & 60.0 & & 1419.8 & 3754.9 & 8369.9 \\
    \hline
    Qwen + R1 & {93.6} & {89.2} & {66.7} & & 662.2 & 2656.4 & 6966.8 \\
    Qwen + QwQ & 96.0 & 90.4 & 56.7 & & 1424.1 & 3754.3 & 8512.4 \\
    QwQ + R1 & 47.7 & 55.4 & 50.0 & & 495.4 & 2288.3 & 7207.6\\
    Qwen + R1 + QwQ & 94.9 & 89.6 & 40.0 & & 878.8 & 2558.5 & 7767.7 \\
    \bottomrule[0.8pt]
    \end{tabular}
    \caption{Evaluations of average merging on Qwen-32B models.}
    \label{tab:32B_res_qwq}
\end{table}

Additionally, we conducted some preliminary experiments to merge QwQ-32B \citep{qwq}, another variant of a System 2 reasoning model derived from Qwen2.5-32B. Using average merging, we observed, as shown in Table \ref{tab:32B_res_qwq}, that the merged model based on QwQ-32B tends to produce excessively lengthy responses, regardless of whether the other merging model is Qwen-32B or R1-32B. The worst performance occurs when directly merging two System 2 reasoning models, i.e., QwQ-32B and R1-32B, which highlights the significant parameter disparity between these two models. This disparity likely stems from their differing training strategies, as R1-32B is obtained through direct fine-tuning, whereas QwQ-32B is trained via reinforcement learning. However, when an intermediate base model is introduced into the merging process, the resulting merged model (Qwen+R1+QwQ) performs comparably, underscoring the critical role of a pre-trained model in facilitating effective model merging.

\begin{mybox}[colback=gray!10]{Takeaway 5}
\textbf{\textit{The merging of large-scale models poses significant challenges in simultaneously maintaining reasoning performance while substantially reducing response length. The substantial performance gaps between the merging models likely contribute to this difficulty.}}
\end{mybox}
\section{Further Analysis} \label{sec:analysis}
In this section, we further investigate some interesting questions, such as whether the merged model retains the ability for self-critic and self-correction, how the response length correlates with the difficulty of the questions, the special characteristics of model merging on long-to-short reasoning task, discussion about the relations between long-to-short and short-to-long, and some of our failure experience.

\begin{figure}[t]
    \centering
    \begin{subfigure}[b]{0.39\textwidth}
        \centering
        \includegraphics[width=\textwidth]{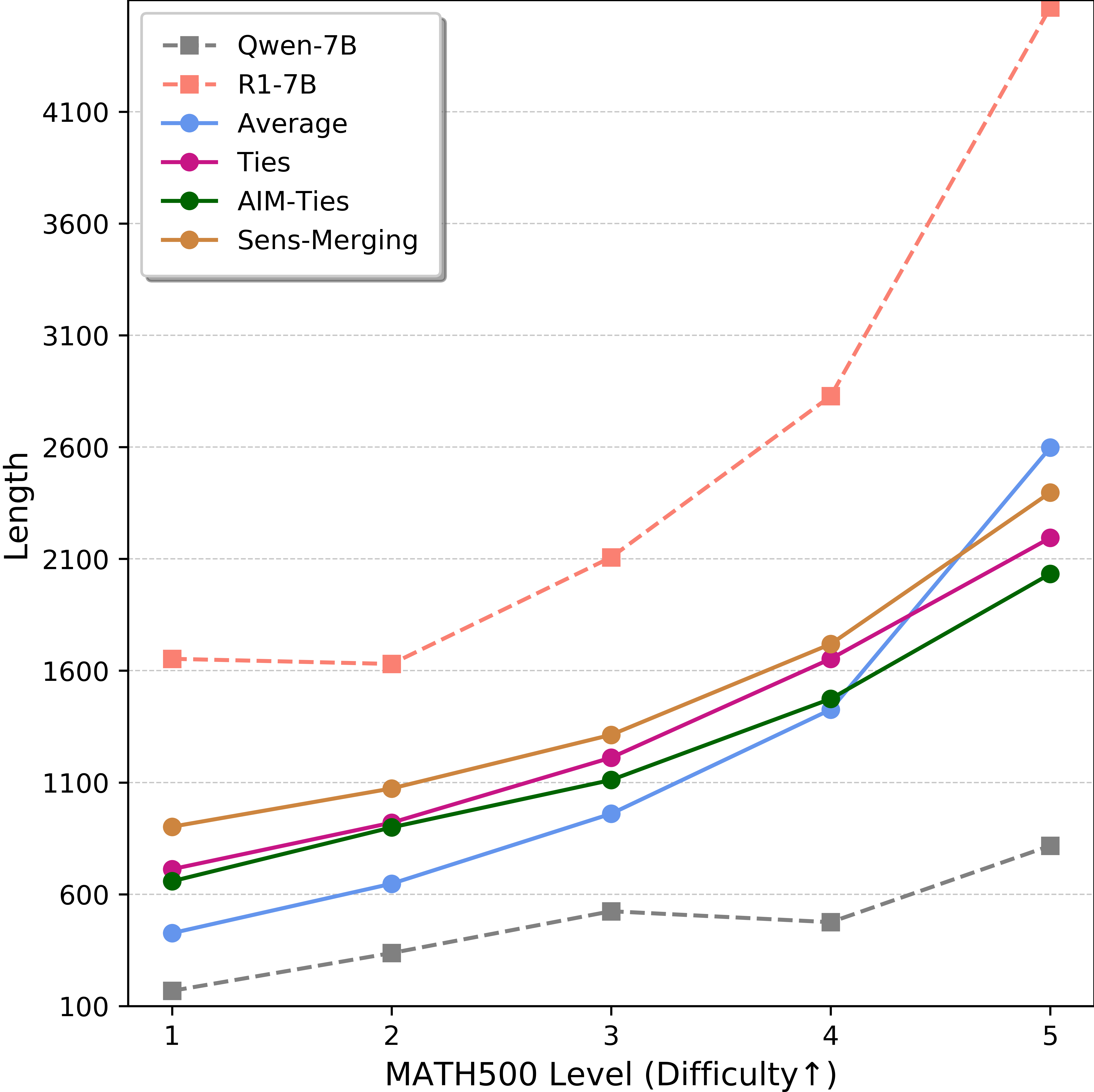}
        \caption{Response length VS. Difficulty}
    \end{subfigure}
    \hfill
    \begin{subfigure}[b]{0.58\textwidth}
        \centering
        \includegraphics[width=\textwidth]{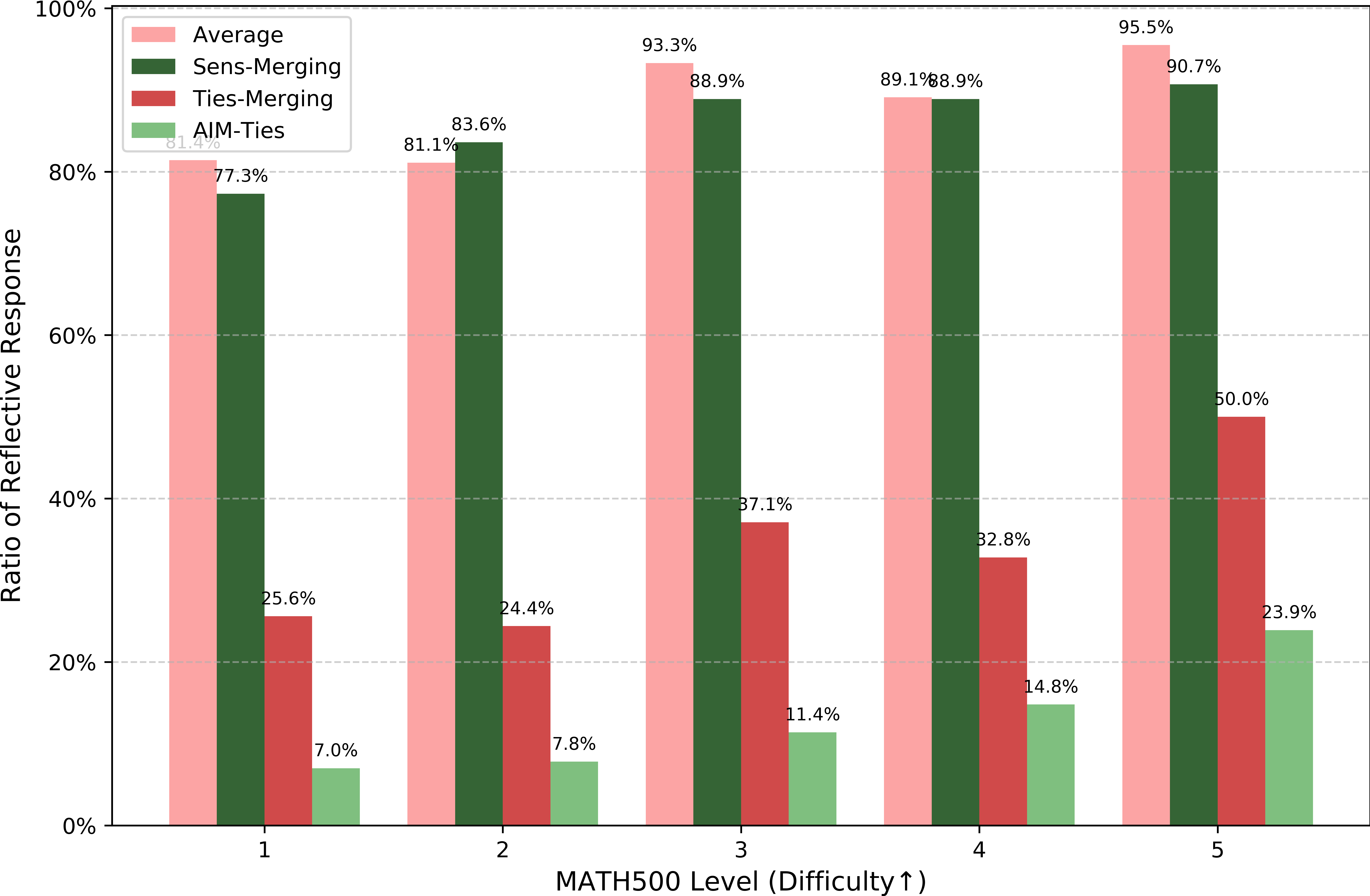}
        \caption{Reflection VS. Difficulty}
    \end{subfigure}
    \caption{Changes in response length and the ratios of reflective responses corresponding to different difficulty levels on the Math500 dataset.}
    \label{fig:difficulty_level}
\end{figure}

\paragraph{How the response length correlates to the difficulty of the question?}
From the perspective of datasets, as shown in Table \ref{tab:7b_merge}, longer responses are generated for more challenging datasets. This trend is consistent across the System 1 quick-thinking model, the System 2 reasoning model, and the merged models. Specifically, when examining the MATH500 dataset with varying difficulty levels, the same conclusion can be drawn: response length positively correlates with the difficulty level of the questions, as shown in Figure \ref{fig:difficulty_level}(a).

\paragraph{Whether the merged model retains the ability for self-critique and self-correction?}
To address this question, we calculate the ratios of reflective responses. A response is defined as reflective if it contains any of the pre-defined keywords\footnote{The keywords include: wait, re-examine, recap, double-check, let me (just) check, and let me (just) verify.}. Our observations indicate that approximately 99.3\% of the responses generated by the DeepSeek-R1-7B model are reflective, whereas the corresponding ratio for the Qwen2.5-Math-7B model is nearly 0. This result highlights both the accuracy and coverage of the selected keywords.

\begin{table}[ht]
    \centering
    \def\arraystretch{1,2}
    \begin{tabular}{cccccccc}
    \toprule[0.8pt]
       \multirow{2}{*}{\diagbox{Method}{Bench}} & GSM8K & MATH500 & \makecell[c]{Minerva\\Math} & \makecell[c]{Olympiad\\Bench} & \makecell[c]{College\\Math} & AIME24 & \multirow{2}{*}{Avg.} \\
       & (1319) & (500) & (272) & (675) & (2818) & (30) \\
       \hline
        DeepSeek-R1-7B & 98.3 & 99.4 & 98.2 & 99.4 & 99.5 & 100 & 99.1 \\
        DPO & 98.3 & 99.4 & 99.6 & 99.4 & 99.2 & 100 & 99.3\\
        \hline
        Average Merging & 27.3 & 89.6 & 97.1 & 97.8 & 94.7 & 100 & 84.4\\
        Task Arithmetic & 56.7 & 83.6 & 73.9 & 81.2 & 81.7 & 76.7 & 75.6\\
        TIES-Merging & 19.2 & 36.2 & 16.9 & 42.4 & 18.2 & 50.0 & 30.5\\
        LoRE-Merging & 42.0 & 92.6 & 93.8 & 95.9 & 94.0 & 96.7 & 85.8\\
        AIM-TIES & 21.7 & 40.3 & 21.7 & 47.7 & 31.0 & 93.3 & 42.6\\
        Sens-Merging & 50.2 & 83.6 & 72.4 & 78.7 & 78.7 & 76.7 & 73.4\\
    \bottomrule[0.8pt]
    \end{tabular}
    \caption{Ratios (\%) of responses containing reflective content across various datasets. Scores for Qwen2.5-Math-7B are not reported, as it produces almost no reflective responses.}
    \label{tab:reflection_rates}
\end{table}

As shown in Table \ref{tab:reflection_rates}, all models produced through various model merging methods exhibit the ability for self-critique and self-correction. Several observations can be made:
1) The reflection ratio does not correlate with reasoning accuracy. For instance, both TIES-Merging and AIM-TIES achieve excellent performance, yet their reflection ratios on simpler tasks, such as GSM8K (19.2\%; 21.7\%) and MATH500 (36.2\%; 40.3\%), remain relatively low. Conversely, Average Merging and LoRE-Merging achieve over 80\% reflective responses while they exhibit inferior performance.
2) The reflection ratio is positively associated with response length, which aligns with an intuitive expectation.
3) As shown in Figure \ref{fig:difficulty_level}(b), with the increasing of difficulty level, more reflective responses are generated.

Based on these observations, we hypothesize how the merged models maintain reasoning accuracy while reducing response length. For merged models with a high proportion of reflective responses, such as LoRE-Merging and Sens-Merging, the reduction in response length is primarily achieved by eliminating redundant reasoning content and streamlining the reasoning process, while still performing reflections in most cases. This assumption is supported by the statistics on the MATH500 dataset, which indicate that the frequency of reflection keywords appearing in each response of Sens-Merging and LoRE-Merging is 2.5, compared to 11.1 per response of the DeepSeek-R1-7B model.
In contrast, merged models like TIES-Merging and AIM-TIES inherently possess the ability to quickly respond to simpler questions and engage in more deliberate reasoning for complex ones, as evidenced by the distribution of reflective responses across various datasets.

\paragraph{Special Characteristics of Model Merging on Long-to-Short Reasoning} In traditional model merging, the merging objectives can be categorized into two main branches: merging models from different domains into a unified model to integrate multi-domain capabilities, and merging checkpoints from the training process to enhance the final performance on a single task. Unlike multi-domain model merging, which can typically decompose domain-specific knowledge from the task model, identifying feature vectors that represent quick-thinking or slow-thinking capacities poses a significant challenge. Moreover, the models being merged are often general-purpose models and are evaluated on general reasoning benchmarks rather than domain-specific datasets. In contrast to checkpoint merging, where the candidate models exhibit minimal parameter shifts, long-to-short merging involves models with substantial parameter differences, as well as variations in response style and downstream task performance, making the merging considerably more challenging.

\paragraph{From Long-to-Short to Short-to-Long} Unlike the training-based methods \citep{ma2025cotvalvelengthcompressiblechainofthoughttuning, xia2025tokenskipcontrollablechainofthoughtcompression} which need to change the base model for training, model merging offers a more straightforward approach to achieving short-to-long adjustments on quick-thinking models by simply tuning the merging weights or selecting an appropriate base merging model, without the need for further training. For instance, in a simple attempt to achieve short-to-long reasoning, we assigned a small weight (e.g., 0.2) to the slow-thinking model while using the quick-thinking model as the base. This approach resulted in an approximate improvement of over 20 points in the overall score, accompanied by a 25\%+ increase in response length. Model merging provides a more effective and efficient solution for enabling System 1 models to acquire System 2 reasoning abilities compared to model distillation \citep{yu2024distilling21}.

\paragraph{Failure Experience and Future Directions}
Apart from the successful attempts, we also encountered numerous failure cases in this study. Here, we share some key insights from these critical failures and provide potential future directions in this field.

\underline{\textit{Sensitivity of merging hyper-parameters.}} Most merging methods are sensitive to hyper-parameters. For instance, task-vector-based merging methods involve a coefficient hyperparameter $\alpha \in [0, 1]$, which determines the contribution proportions of different merging models to the final model. To preserve overall reasoning accuracy, $\alpha$ is typically set to $\alpha > 0.5$, indicating that the System 2 model plays a more significant role. As expected, we observe a positive correlation between both reasoning accuracy and response length with increasing $\alpha$. However, we regretfully find that even small variations in $\alpha$ (e.g., within a range of 0.1) result in a moderate performance gap (e.g., around 0.5-1.0 points on overall performance), highlighting a sensitivity issue.
The sensitivity characteristics imposes additional experimental burdens to identify the optimal merging settings. Therefore, it is highly desirable to develop a hyperparameter-insensitive merging method or a framework capable of automatically determining the optimal parameters.

\underline{\textit{Sensitivity to calibration data in activation-based merging methods.}}
Activation-based model merging methods demonstrate remarkable performance in maintaining reasoning accuracy while achieving response length reduction. However, we observe that the choice of calibration data significantly influences the merging performance. For example, when sampling calibration data from the s1K dataset, which contains aligned short and long CoT data for each question, different methods exhibit varying preferences. AIM-TIES performs better when using the question paired with its short answer as activations, whereas Sense-Merging achieves superior results when incorporating both the short and long answers along with the question. Additionally, the sample size of calibration data also impacts merging effectiveness. In our experiments, we tested calibration with only 10 and 20 data samples, observing a performance gap of approximately 1.0 point in the average score. Further exploration of calibration data selection is left for future work.

\underline{\textit{Challenging of merging models having large performance dispensary.}}
The moderate merging performance observed on 14B and 32B models indicates that not all quick- and slow-thinking model pairs can be effectively merged using existing model merging methods, particularly when evaluated on tasks where the candidate models exhibit significant performance disparities. In this context, exploring strategies to enable the merged model to compare with even surpass the upper bound of the superior model with shorter response lengths, especially when there is a substantial performance gap between the candidate models, presents an intriguing research direction.

\section{Conclusion}

Our study demonstrates model merging effectively addresses Long-to-Short reasoning in LLMs by integrating System 1's efficiency with System 2’s rigor, reducing response lengths by up to 55\% without compromising performance. Task-vector merging balanced simplicity and effectiveness, while activation-informed methods showed promising improvement potential. These findings position model merging as a cost-efficient solution to mitigate overthinking while preserving reasoning quality. Future work should explore theoretical foundations and refine merging strategies to broaden real-world applicability.

\bibliography{iclr2025_conference}
\bibliographystyle{iclr2025_conference}

\appendix
\section{Appendix}
\subsection{Experiments Configuration}
\paragraph{DPO Training} We use the full set of s1K dataset for the DPO training. Due to the limited computational resource, we conducted the LoRA training with lora rank of 16. The maximum length is set to 4096 and the learning rate is 5e-5. We train the dataset for 3 epoch and report the best performance of the saved checkpoints.

\paragraph{Configurations} We report our merging configurations on various merging methods as Table \ref{tab:config}. All merged models are saved and evaluated in BF16, except for those produced by Sens-Merging, as the sensitivity factors are calculated on the CPUs, which does not support BF16 computations. To preserve the accuracy of the calculated sensitivity scores, the merged models generated through Sens-Merging are saved in FP32.
For evaluations, we use the standard Qwen evaluation toolkit. For System 1 models, we use the `cot' prompt with few-shots demonstrations and a maximum length of 8192. For System 2 models, we use `qwen25-math-cot' prompt in a zero-shot setting with a maximum length of 10240.
We also conducted experiments on LLaMA3.1-8B models; however, we were unable to reproduce the reasoning accuracy reported in their work using the Qwen evaluation toolkit. As a result, we do not report the results for the LLaMA-series models.

\begin{table}[ht]
    \centering
    \def\arraystretch{1,2}
    \begin{tabular}{c|cccc}
    \toprule[0.8pt]
    \multirow{2}{*}{Method} & \multicolumn{4}{c}{Hyper-parameters} \\
    \cline{2-5}
    & 1.5B & 7B & 14B & 32B \\
    \hline
    Task Arithmetic & $\alpha=0.7$ & $\alpha=0.7$ & $\alpha=0.7$ & $\alpha=0.7$ \\
    TIES-Merging & $k=0.8$, $\alpha=1.0$ & $k=0.8$, $\alpha=1.0$ & $k=0.2$, $\alpha=0.5$ & $k=0.25$, $\alpha=0.55$ \\
    DARE & $p=0.3$ & $p=0.3$ & $p=0.4$ & - \\
    AIM-TIES & $\omega=0.4$ & $\omega=0.4$ & $\omega=0.4$ & - \\
    Sens-Merging & $\alpha=0.4$, $T=3.0$ & $\alpha=0.7$, $T=2.0$ & $\alpha=0.8$, $T=6.0$ & - \\
    \bottomrule[0.8pt]
    \end{tabular}
    \caption{The configurations of various merging methods. $\alpha$ means the coefficient in TA merging. $p$ means the drop rate in DARE merging. $k$ denotes the trim ratio in TIES-Merging. $\omega$ means the balance factor in AIM. $T$ is the temperature in Sens-Merging.}
    \label{tab:config}
\end{table}

\end{document}